\documentclass[notoc]{kling-keye}
\usepackage{multirow,colortbl,pifont,xspace,enumitem,float,calc,listings}
\lstdefinestyle{prompt}{basicstyle=\ttfamily\footnotesize,columns=fullflexible,breaklines=true,breakatwhitespace=true,keepspaces=true,showstringspaces=false,aboveskip=6pt,belowskip=6pt}

\newcommand{\harness}{\textsc{video-relook}\xspace}
\newcommand{\mme}{Video-MME-v2\xspace}

\newcommand{\abench}{FastAct\xspace}

\title{Rethinking Video Agents: When and Why Video Harnesses Work}
\newcommand{\paperauthors}{%
 {\normalfont\begin{tabular}{@{}c@{\hspace{2em}}c@{\hspace{2em}}c@{}}
 Luoyuan Zhang\textsuperscript{*,\textsection} & Xuanyu Zheng\textsuperscript{*,\textdagger,\textdaggerdbl} & Yancheng Long\textsuperscript{*} \\[3pt]
 Haoyang Xu\textsuperscript{*} & Bin Wen\textsuperscript{\textdagger} & Tingting Gao
 \end{tabular}}%
}
\author{\paperauthors}
\newcommand{\authornotes}{%
 \begingroup
 \renewcommand{\thefootnote}{\fnsymbol{footnote}}%
 \footnotetext[1]{Equal contribution.}%
 \footnotetext[2]{Corresponding authors.}%
 \footnotetext[3]{Project lead.}%
 \footnotetext[4]{This work was conducted during the author's internship at KlingAI.}%
 \endgroup
}
\affiliation{KlingAI Research}
\headertext{Rethinking Video Agents}
\reportdate{November 18, 2026}
\hypersetup{pdftitle={Rethinking Video Agents: When and Why Video Harnesses Work},pdfauthor={Luoyuan Zhang, Xuanyu Zheng, Yancheng Long, Haoyang Xu, Bin Wen, Tingting Gao}}

\projectlinks{}

\newcommand{\teasercaption}{\textbf{From overview to confirmation.} The example illustrates how an agentic framework helps the model understand the details of the scene.}

\reportabstract{

Video understanding depends not only on how models reason about visual evidence, but also on which evidence they observe. Fixed observation policies determine this evidence in advance; agentic harnesses allow models to seek further observations as their understanding develops. This shift creates an opportunity to realize capabilities that fixed inputs leave untapped, yet the ability to request observations does not ensure their effective use. We investigate when and why video harnesses improve understanding, examining how task demands and models' capacity for interaction jointly shape their benefits. Using Video-Relook as an experimental framework, we let models adapt their temporal and spatial observations from a common default configuration, and analyze performance and interaction trajectories across models and tasks. Adaptive observation yields substantial gains on fine-grained action understanding, with benefits also extending to Video-MME-v2. Even under the same observation interface, gains differ substantially across models; interaction traces reveal corresponding differences in evidence seeking and the reliability of answer revision. The benefits also vary across tasks for the same model, ranging from substantial improvements to limited gains and regressions. Together, these findings support an account in which harness effectiveness depends on the alignment between opportunities for informative observation and the model's ability to turn interaction into reliable revisions. They motivate video harnesses that coordinate perception and reasoning, matching observational autonomy to task demands and model capabilities.

}
\begin{document}
\maketitle
\authornotes
\reportcontents
\begin{figure}[H]
\centering
\includegraphics[width=\linewidth]{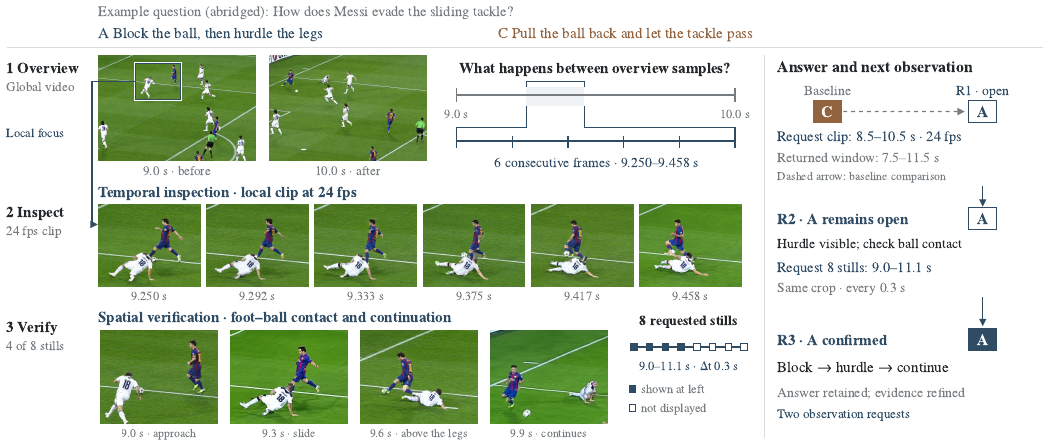}
\caption{\teasercaption}\label{fig:teaser}
\end{figure}
\section{Introduction}
\label{sec:intro}
The agentic paradigm is extending into visual understanding, allowing models to participate in deciding how they acquire and process visual information. For video tasks, this shift can be implemented through tools for extracting clips, cropping spatial regions, and selecting frames, together with a loop that organizes their use. Yet these components define what a model can do, rather than explain why those operations should improve understanding. A system capable of calling visual tools must still decide when to look again, which observations could resolve its uncertainty, and whether new evidence warrants revising an answer. Bringing the agentic paradigm to video therefore raises a question beyond tool availability: \emph{under what conditions does autonomous control over observation translate into better video understanding?}

This question arises from the interdependence of perception and reasoning. A model's judgments depend on the visual evidence available to it, while uncertainty arising during reasoning determines which additional evidence is worth seeking. Fixed observation policies constrain this process in advance: the model must interpret the video through a prescribed temporal and spatial sampling scheme, which may not present relevant details clearly enough. Autonomous observation instead allows the model to adapt how it acquires evidence to the question and its developing interpretation. Our starting point is a common default observation configuration, from which a suitable policy can emerge through interaction, reducing reliance on manually chosen benchmark-specific settings. This creates an opportunity to realize capabilities that fixed inputs leave untapped, while placing new demands on the model: it must recognize unresolved questions, organize informative observations, and turn the returned information into reliable judgments.

Fine-grained action understanding provides a concrete setting in which to study this relationship. Actions with similar overall semantics may yield different answers because of a brief contact, a local pose, or the order of events. Recognizing that a player evades a sliding tackle does not establish how he controls the ball or when he hurdles the defender's legs. An overview may preserve the gist of the action without resolving the details that determine the answer, giving densely sampled local clips and spatial detail a clear role. At the same time, more frames cannot substitute for correctly interpreting contact relations and action order. This setting exposes both the opportunity for autonomous observation and the capabilities it requires: we can examine whether models actively seek relevant evidence and whether their judgments improve as observations accumulate. High temporal resolution is thus one condition that makes the problem visible; its broader relevance requires evaluation across video tasks with different evidence demands.

We study these conditions using \harness as an experimental framework. Starting from a default global overview, the model requests local clips, selected frames, or spatial detail according to its current understanding, receives observations, and revises its answers within a bounded multi-turn conversation. Our study varies two dimensions: different models use a common observation interface and round budget, and the same model encounters different types of video tasks. We retain first-round predictions, answer trajectories, observation requests, and stopping records to trace gains and losses.

The results first establish the scope for improvement through adaptive observation. All five evaluated models obtain positive gains on fine-grained action understanding, reaching 14.2 percentage points; on \mme, Gemini 3.8 Flash, Gemini 3.6 Flash, and Qwen 3.8 Flash gain 3.78, 1.91, and 5.25 points, respectively, whereas Qwen 3.7 Flash shows no net gain. Yet greater baseline headroom does not automatically yield greater gains. Qwen 3.7 Flash leaves the most errors to correct but improves by only 3.0 points and stops after the first round on 55.7\% of question groups; Qwen 3.8 Flash and Gemini 3.6 Flash improve by 12.7 and 14.2 points. Together with differences in error recovery and correct-answer retention, these results show that available headroom must be realized through the model's observation and revision behavior. Cross-task results further reveal the conditions on this process: models that benefit substantially on action questions can obtain limited gains or regress elsewhere. For example, Gemini 3.6 Flash loses 1.20 points on TempCompass but gains 2.73 on Video-MMMU multiple-choice questions. The same model--harness pair produces different outcomes across tasks, indicating that observational autonomy also depends on whether further observation can address the difficulties a task presents.

Our contributions are threefold:
\begin{enumerate}
    \item We hypothesize that video harnesses work when informative observation opportunities align with a model's ability to acquire evidence and reliably revise its judgments.
    \item We introduce \harness, a framework for adaptive temporal and spatial observation from a shared default.
    \item We characterize the benefits and limits of autonomous observation across models and tasks, with gains of up to 14.2 percentage points.
\end{enumerate}

\section{Related Work}
\label{sec:related}
\paragraph{Video understanding and its evidence requirements.}
Action datasets~\cite{goyal2017something,shao2020finegym} distinguish fine-grained motion; NExT-QA~\cite{xiao2021nextqa} and MVBench~\cite{li2024mvbench} probe temporal reasoning; Video-MME~\cite{fu2025videomme} spans durations and tasks. \mme~\cite{mmeteam2026videommev2} adds reasoning levels; Video-MMMU~\cite{hu2025videommmu} studies learning from professional videos. Moment-Video~\cite{momentvideo2026} finds errors even when brief events are covered.

\paragraph{Adaptive access to video.}
Active perception~\cite{bajcsy1988active,ballard1991animate} and recurrent attention~\cite{mnih2014attention} treat observation as a decision. VideoAgent~\cite{wang2024videoagent,fan2024videoagent} and VideoTree~\cite{wang2025videotree} adapt access to the question. EVA~\cite{zhang2026eva} learns temporal and spatial selection through staged fine-tuning and reinforcement learning. VideoNexus~\cite{long2026videonexusvideoperceptionreasoning} separates video perception from planning through compact evidence exchange. We compare fixed models under a shared interface.

\paragraph{Visual tools for reasoning.}
VisProg~\cite{gupta2023visprog} and ViperGPT~\cite{suris2023vipergpt} compose tools through programs; $V^*$~\cite{wu2024vstar} guides visual search, and Visual Sketchpad~\cite{hu2024visualsketchpad} supports visual intermediate reasoning. \harness provides clips, crops, and stills during inference.

\paragraph{Evaluating interactive video systems.}
For language and multimodal models~\cite{yu2026minicpm,cui2026minicpm,zhang2026liberating,yao2025minicpm,bai2025qwen3vl,xiaomi2025mimovl,huang2026step3vl,kwaikeyeteam2026kwaikeyevl20technicalreport}, ReAct~\cite{yao2023react} interleaves reasoning and action; Reflexion~\cite{shinn2023reflexion} and Self-Refine~\cite{madaan2023selfrefine} use feedback for revision. Self-correction without external feedback can also degrade reasoning~\cite{huang2024selfcorrect}. AgentVidBench~\cite{agentvidbench2026}, Video-Arena~\cite{videoarena2026}, and VideoGAIA~\cite{videogaia2026} evaluate interactive video understanding. We trace gains and losses through observation requests, stopping, and revision, distinguishing continued interaction from reliable use of newly acquired visual evidence.

\section{Research Framework and Design}
\label{sec:protocol}
\harness provides a common observation interface for comparing models and tasks. Round-level answers, requests, and stopping decisions link final accuracy to the model's interaction history.

\subsection{From fixed to autonomous observation}
The single-pass baseline answers from a prescribed global overview. The harness starts with the same overview request and lets the model seek local temporal and spatial observations, interpret them in the same conversation, and revise its answers. The model and scoring rules remain fixed.

The comparison evaluates the complete harness---prompts, observation decisions, and multi-turn inference---without matching compute or isolating visual input. We separate revisions preceding new observations from those following their return.

\begin{figure}[htbp]
\centering\includegraphics[width=\textwidth]{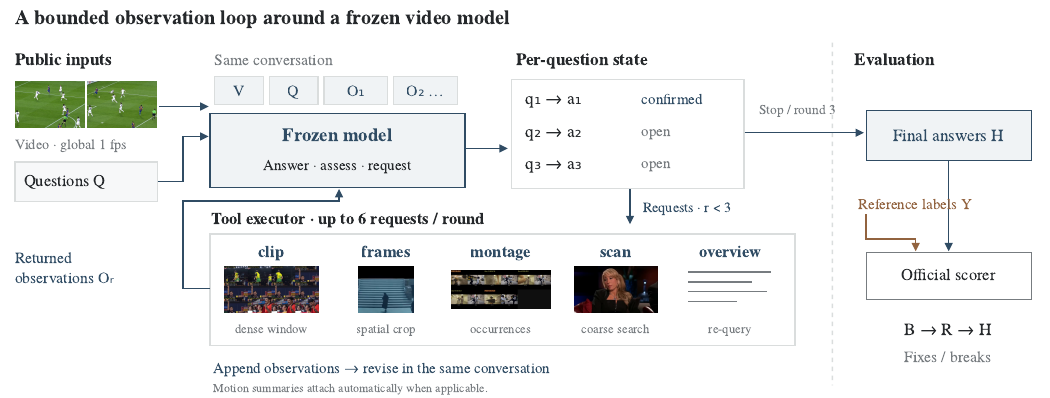}
\caption{Requests select observations; the return loop appends them to the same conversation. Reference labels enter only the scorer.}
\label{fig:framework}
\end{figure}

\subsection{A bounded observation loop}
\label{sec:protocol_rounds}
The model receives the overview and questions, produces provisional answers with evidence notes, and may request observations. Each request specifies a question, tool, and parameters. The executor returns source-video observations to the same conversation. All session evaluations allow at most three model rounds and six requests per request-producing round; round three returns final answers.

\paragraph{Overview and request generation.}
The overview requests 1\,fps global sampling; provider preprocessing is recorded separately, as APIs do not expose their internal visual representation. The first-round prompt asks for evidence before answering, marks unresolved questions, and supplies checklists inferred from the question text. Reference answers, benchmark tags, and separate evidence annotations are withheld; time ranges in questions remain visible. Different prompts can yield different baseline and first-round answers.

\paragraph{Observation interface.}
\texttt{clip} revisits a window at up to 24\,fps with optional spatial cropping; \texttt{frames} returns labelled stills and optional enlarged crops; \texttt{montage} groups candidate occurrences with boundary context for counting; \texttt{scan} coarsely searches a longer interval; and \texttt{overview} re-queries the global input. Selected motion questions also receive a motion-energy summary. Implementation details appear in the supplementary material.

\paragraph{Revision and stopping.}
Later rounds append observations and allow answers, further requests, or stopping. Confirmed answers are retained for questions receiving no observation of their own. The loop ends when answers are available without further requests or the round limit is reached. If unanswered questions receive no initial requests, the executor can check a default window.

\subsection{Measuring gains and tracing revisions}

Paired answers identify the corrections and regressions behind the accuracy gain; first-round predictions locate these changes before or after new observations.

Let $B_i$ and $H_i$ indicate whether the single-pass baseline and the harness answer question $i$ correctly. On the same $N$ questions, define a \emph{fix} as $(B_i,H_i)=(0,1)$ and a \emph{break} as $(1,0)$. The accuracy difference is exactly
\begin{equation}
\Delta A = \frac{N_{\mathrm{fix}}-N_{\mathrm{break}}}{N}.
\label{eq:net}
\end{equation}
To account for different baseline accuracies, define recovery $r$ over baseline errors and regression $d$ over correct answers:
\begin{equation}
 r=\Pr(H=1\mid B=0),\quad d=\Pr(H=0\mid B=1).
\label{eq:rates}
\end{equation}
Writing baseline accuracy as $a$, we obtain
\begin{equation}
\Delta A=(1-a)r-ad.
\label{eq:balance}
\end{equation}
For a strong baseline, small regression rates can offset substantially larger recovery rates.

\paragraph{Separating the stages of revision.}
Let $R_i$ denote correctness after the first harness round, before any requested observation is returned. Then
\begin{equation}
\Delta A=\underbrace{(A_R-A_B)}_{\text{initial-round change}}
          +\underbrace{(A_H-A_R)}_{\text{later-round change}}.
\label{eq:stages}
\end{equation}
The first term includes prompting and inference variability; the second includes new observations, further reasoning, and accumulated context. Neither isolates the causal effect of visual evidence. We enumerate $(B_i,R_i,H_i)$ to distinguish persistent early errors from later reversals. Unanswered predictions count as incorrect.

\section{When Do Video Harnesses Work?}
\label{sec:exp}
We compare gains across tasks and models, then examine temporal demands and question categories.

\subsection{Experimental setting}
\label{sec:exp_setup}
\paragraph{Tasks.}
We evaluate \mme~\cite{mmeteam2026videommev2}, TempCompass~\cite{liu2024tempcompass}, Video-Holmes~\cite{cheng2025videoholmes}, and the multiple-choice portion of Video-MMMU~\cite{hu2025videommmu}. For a closer study of action details, we use \abench, an internal benchmark drawn from high-quality video data with human annotation and review. It tests brief contacts, local movements, and action order.

\paragraph{Action details in FastAct.}
\abench contains 1,000 four-option questions over 100 silent clips of approximately 36--60 seconds, covering sports, combat, dance, and other actions. Questions concern contact, object manipulation, action order, and participant reactions. A bucket-kick question, for example, distinguishes the lifting foot, supporting leg, and striking foot. Questions specify the relevant time range; annotated evidence intervals have a median duration of 0.90 seconds. Videos are provided at 24\,fps, with 693 questions annotated as requiring 8\,fps and 307 as requiring 24\,fps. The supplied time ranges reduce the need for broad search, allowing us to examine whether targeted observation resolves details that remain ambiguous in the overview.

\paragraph{Models and comparisons.}
FastAct covers Qwen 3.7 Flash, Qwen 3.8 Flash, Gemini 3.6 Flash, Gemini 3.8 Flash, and Gemini 4 Argon; public benchmarks cover the four Flash models. Each model is compared with its single-pass baseline, holding weights, options, and scoring fixed. Both arms request a 1\,fps overview; the harness adds local observations. Gemini uses HIGH thinking; Qwen uses the gateway default.

\begin{figure}[htbp]
\centering
\includegraphics[width=\textwidth]{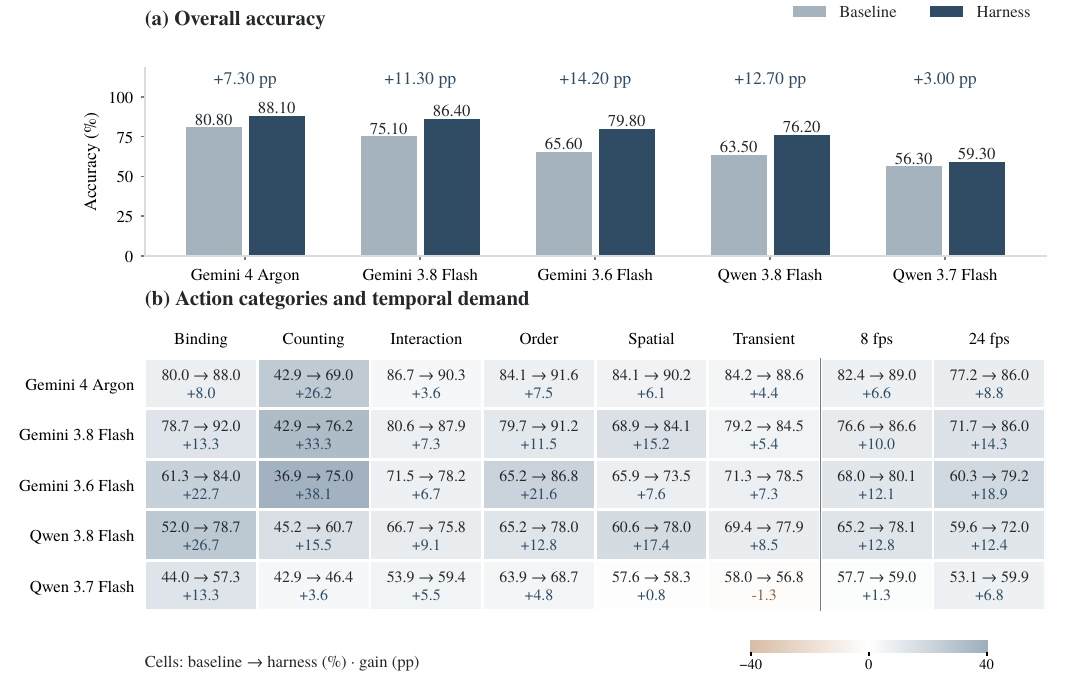}
\caption{FastAct overall and subgroup results. Paired bars compare baseline and harness accuracy; cells report paired accuracies and gains, with shading encoding the gain.}
\label{fig:fastact_results}\label{fig:action_profiles_main}
\end{figure}

\subsection{A revealing setting for adaptive observation}
\label{sec:transfer}
An overview may miss a decisive detail, or the model may misinterpret available evidence. Fine-grained actions expose both difficulties: a brief contact or change in limb position can distinguish answers within an easily recognized activity.

FastAct gives the relevant time range but leaves the action details to be resolved. Dense clips and spatial crops permit inspection; the model must decide what to request and whether the evidence warrants revision. This couples an opportunity to recover overlooked details with the need to direct and interpret observation.

\begin{figure}[htbp]
\centering
\includegraphics[width=\textwidth]{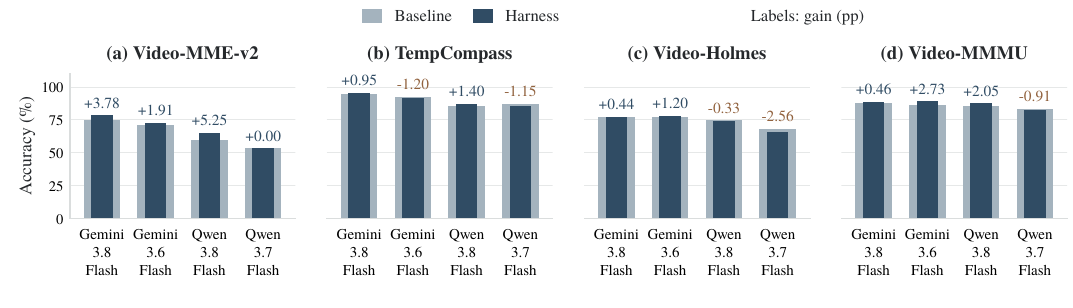}
\caption{Accuracy across four public benchmarks. Harness bars overlay the baseline; labels show gains in percentage points.}
\label{fig:benchmark_comparison}
\end{figure}

Public benchmarks show the limits of this benefit (\cref{fig:benchmark_comparison}): Gemini 3.6 Flash gains 2.73 points on Video-MMMU but loses 1.20 on TempCompass; Qwen 3.8 Flash gains 5.25 on \mme but loses 0.33 on Video-Holmes. We examine the associated interaction behaviors below.

\subsection{Gains from autonomous observation}
\label{sec:exp_public}
FastAct gains range from 3.0 to 14.2 points across the five models (\cref{fig:fastact_results}). Qwen 3.7 Flash's 95\% interval spans zero ($-0.4$ to $6.4$ points), leaving its improvement less certain.

\paragraph{Diverse video understanding.}
On \mme, Gemini 3.8 Flash, Gemini 3.6 Flash, and Qwen 3.8 Flash gain 3.78, 1.91, and 5.25 points, respectively, while Qwen 3.7 Flash remains at 53.16\% (\cref{fig:benchmark_comparison}). In the same model order, official group scores change from 62.21 to 65.95, 56.30 to 58.45, 43.80 to 49.78, and 36.32 to 36.46. The benefit extends beyond action understanding.

\subsection{Headroom must be realized through interaction}
On FastAct, Qwen 3.7 Flash has the weakest baseline and smallest gain. Qwen 3.8 Flash and Gemini 3.6 Flash start higher yet improve more; even the strongest baseline, Gemini 4 Argon, gains 7.3 points. Baseline headroom alone does not explain the ranking of gains.

\begin{figure}[htbp]
\centering
\includegraphics[width=\textwidth]{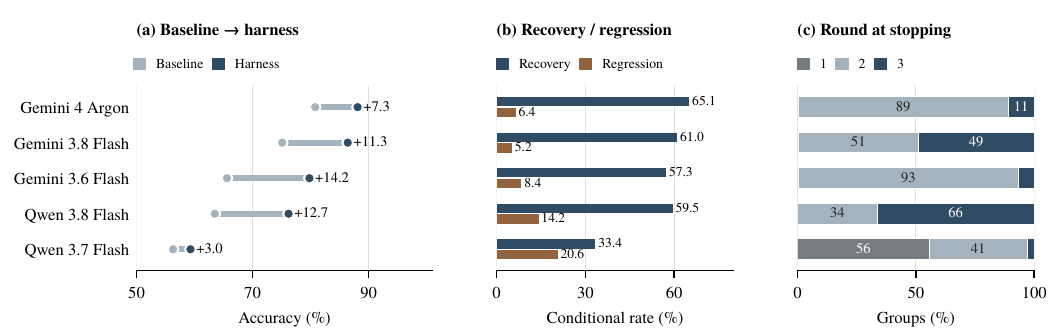}
\caption{\textbf{Accuracy, revision, and stopping on FastAct.} Recovery is conditional on baseline errors; regression on baseline successes. Round counts refer to question groups.}
\label{fig:utilization}
\end{figure}

We relate these differences to recovery, regression, and stopping (\cref{fig:utilization,sec:analysis}). A lower baseline leaves more errors available for correction, but says little about whether further observation will help the model resolve them without overturning correct answers.

\subsection{Temporal demand conditions the opportunity}
\label{sec:exp_abench}
\begin{figure}[htbp]\centering
\includegraphics[width=\textwidth]{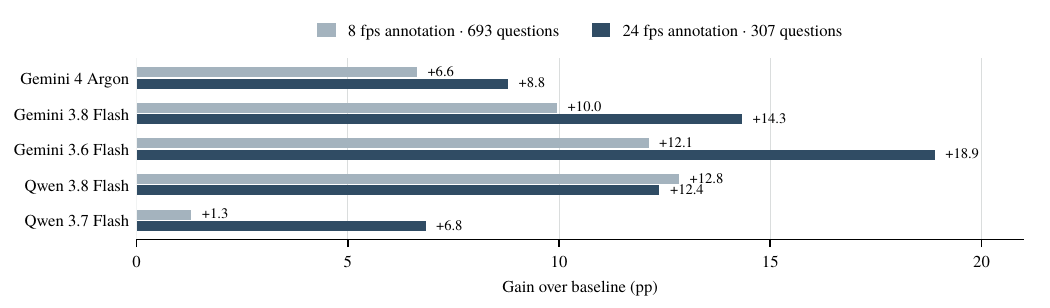}
\caption{\textbf{Gain by annotated temporal demand.} Bars compare the five models within each stratum; these annotations are not a controlled sampling intervention.}
\label{fig:minfps}\end{figure}
\cref{fig:minfps} compares the two temporal-demand strata. Gemini 3.6 Flash gains 12.1 points on the 8\,fps questions and 18.9 on the 24\,fps questions; Gemini 3.8 Flash gains 10.0 and 14.3. Qwen 3.7 Flash also gains more on the 24\,fps subset, 6.8 points versus 1.3. Qwen 3.8 Flash, however, gains similarly on the two subsets (12.8 and 12.4 points). Higher temporal demand is thus associated with larger gains for several models, but does not determine the gain by itself.

\subsection{Observation opportunities within tasks}
\label{sec:subgroups}
\paragraph{MME categories and cognitive levels}
\begin{table}[htbp]\centering\small
\setlength{\tabcolsep}{5pt}
\caption{Video-MME-v2 accuracy (\%) by category, cognitive level, and group type. Each cell reports baseline $\rightarrow$ harness.}
\label{tab:exp_mme}
\begin{tabular}{@{}lrcccc@{}}\toprule
Subset & N & Gemini 3.8 Flash & Gemini 3.6 Flash & Qwen 3.8 Flash & Qwen 3.7 Flash \\ \midrule
Overall & 3200 & $74.56 \rightarrow 78.34$ & $70.84 \rightarrow 72.75$ & $59.88 \rightarrow 65.13$ & $53.16 \rightarrow 53.16$ \\
\midrule
Action \& Motion & 282 & $54.96 \rightarrow 59.22$ & $49.65 \rightarrow 55.32$ & $41.49 \rightarrow 50.71$ & $34.40 \rightarrow 31.91$ \\
Change & 242 & $76.03 \rightarrow 76.86$ & $67.77 \rightarrow 70.66$ & $61.98 \rightarrow 71.90$ & $56.20 \rightarrow 59.92$ \\
Complex Plot Comprehension & 249 & $78.71 \rightarrow 79.52$ & $74.30 \rightarrow 75.10$ & $59.84 \rightarrow 64.26$ & $49.80 \rightarrow 52.61$ \\
Frame-Only & 575 & $78.78 \rightarrow 85.91$ & $78.78 \rightarrow 81.39$ & $76.87 \rightarrow 82.61$ & $73.04 \rightarrow 72.70$ \\
Frames \& Audio & 478 & $88.28 \rightarrow 90.17$ & $89.54 \rightarrow 88.08$ & $62.97 \rightarrow 65.90$ & $52.93 \rightarrow 52.30$ \\
Order & 343 & $84.26 \rightarrow 86.88$ & $82.22 \rightarrow 82.22$ & $72.89 \rightarrow 80.47$ & $72.59 \rightarrow 70.55$ \\
Physical World Reasoning & 298 & $51.34 \rightarrow 65.10$ & $45.97 \rightarrow 45.64$ & $35.23 \rightarrow 41.28$ & $29.53 \rightarrow 26.85$ \\
Social Behavior Analysis & 269 & $67.29 \rightarrow 66.17$ & $61.71 \rightarrow 60.97$ & $46.47 \rightarrow 47.58$ & $39.03 \rightarrow 39.78$ \\
Temporal Reasoning & 182 & $79.12 \rightarrow 80.77$ & $70.88 \rightarrow 74.73$ & $68.13 \rightarrow 67.58$ & $47.80 \rightarrow 53.30$ \\
Video-Based Knowledge Acquisition & 282 & $74.11 \rightarrow 75.89$ & $64.89 \rightarrow 73.40$ & $54.26 \rightarrow 59.22$ & $50.35 \rightarrow 50.00$ \\
\midrule
Level 1 & 1053 & $83.10 \rightarrow 87.84$ & $83.67 \rightarrow 84.43$ & $70.56 \rightarrow 75.02$ & $63.91 \rightarrow 63.44$ \\
Level 2 & 1049 & $73.59 \rightarrow 76.07$ & $68.16 \rightarrow 71.02$ & $61.11 \rightarrow 68.26$ & $54.24 \rightarrow 54.72$ \\
Level 3 & 1098 & $67.30 \rightarrow 71.40$ & $61.11 \rightarrow 63.21$ & $48.45 \rightarrow 52.64$ & $41.80 \rightarrow 41.80$ \\
\midrule
logic & 1124 & $82.12 \rightarrow 84.16$ & $76.60 \rightarrow 78.65$ & $67.62 \rightarrow 71.62$ & $59.96 \rightarrow 60.14$ \\
relevance & 2076 & $70.47 \rightarrow 75.19$ & $67.73 \rightarrow 69.56$ & $55.68 \rightarrow 61.61$ & $49.47 \rightarrow 49.37$ \\
\bottomrule\end{tabular}\end{table}

\cref{tab:exp_mme} reports category results over 800 videos, with failed predictions counted as incorrect. Category and cognitive-level labels are used only for analysis and are not provided to the harness.

Category gains vary by model (\cref{tab:exp_mme}). Qwen 3.8 Flash gains 9.22 points on Action \& Motion but loses 0.55 on Temporal Reasoning. Gemini 3.8 Flash gains 13.76 on Physical World Reasoning but loses 1.12 on Social Behavior Analysis; Gemini 3.6 Flash gains 8.51 on Video-Based Knowledge Acquisition but loses 1.46 on Frames \& Audio. Qwen 3.7 Flash is unchanged overall, yet gains 5.49 on Temporal Reasoning and loses 2.48 on Action \& Motion. Aggregate gains can therefore conceal opposing category changes.

Frame-Only gains extend the benefit beyond temporal integration. Stills and spatial crops are also available, though category scores do not isolate their contributions.

\paragraph{Fine-grained action profiles}
FastAct shows similar variation (\cref{fig:action_profiles_main}): Qwen 3.7 Flash improves on binding but regresses on transient states. Models with larger gains also differ in which categories improve most.

\section{Why Do Video Harnesses Work?}
\label{sec:analysis}
FastAct traces show whether models seek further evidence and whether revisions correct more answers than they break. We also compare final-answer changes on \mme.

\subsection{Seeking evidence and continuing interaction}
On FastAct, Qwen 3.7 Flash ends after one round on 55.7\% of groups, compared with 0--0.3\% for the other four models. It uses 1.47 rounds per group on average, versus 2.66 for Qwen 3.8 Flash and 2.06 for Gemini 3.6 Flash. Qwen 3.7 Flash thus often stops before receiving requested observations. Its stopping behavior describes limited use of this harness, rather than an independent measure of general agentic capability across tools and tasks.

Longer interaction does not rank models by accuracy. Qwen 3.8 Flash reaches round three on 66.3\% of groups, while Gemini 3.6 Flash does so on 6.7\% and achieves higher final accuracy. Round counts measure continuation, not how successfully the returned evidence is used.

\begin{figure}[htbp]
\centering\includegraphics[width=\textwidth]{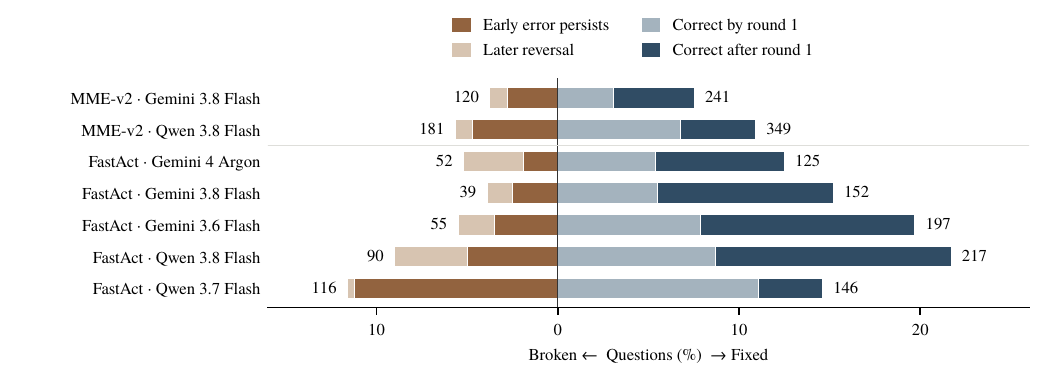}
\caption{\textbf{When answers are fixed or broken.} Bar width is the share of all questions; endpoint labels are counts. Shading separates first-round and later changes.}
\label{fig:flips}\end{figure}

\subsection{Reliable revision determines gains}
On FastAct, Qwen 3.7 Flash corrects 33.4\% of baseline errors but overturns 20.6\% of correct baseline answers (\cref{fig:utilization}). Qwen 3.8 Flash corrects 59.5\% and overturns 14.2\%; Gemini 3.8 Flash corrects 61.0\% and overturns 5.2\%. Both improve error recovery and correct-answer retention.

On \mme, Qwen 3.7 Flash corrects 250 baseline errors but overturns 250 correct answers, leaving its overall accuracy unchanged. Gemini 3.6 Flash corrects 245 and overturns 184, yielding a net gain of 61 answers. Similar numbers of corrections can therefore produce different net gains when correct-answer retention differs.

\cref{fig:flips} separates first-round and later changes for the five FastAct models and for Qwen 3.8 Flash and Gemini 3.8 Flash on \mme. On FastAct, 112 of Qwen 3.7 Flash's 116 baseline-to-final breaks are already incorrect after round one; only four are correct first-round answers overturned later. Most of its regressions therefore precede the return of additional observations.

The losses occur later more often for other models. For Qwen 3.8 Flash, 50 of 90 breaks are persistent first-round errors and 40 are later reversals. For Gemini 4 Argon, later reversals account for 33 of 52 breaks. Later rounds therefore remain a substantial source of regression even for models with larger overall gains.

The same distinction appears in fixes. Of Qwen 3.8 Flash's 217 baseline errors corrected by the final answer, 87 are already correct after round one and 130 become correct afterward. Gemini 3.8 Flash has 55 early and 97 later fixes. These counts classify baseline-to-final changes only: they exclude baseline successes temporarily lost in round one and subsequently recovered.

\begin{figure}[htbp]
\centering
\includegraphics[width=\textwidth]{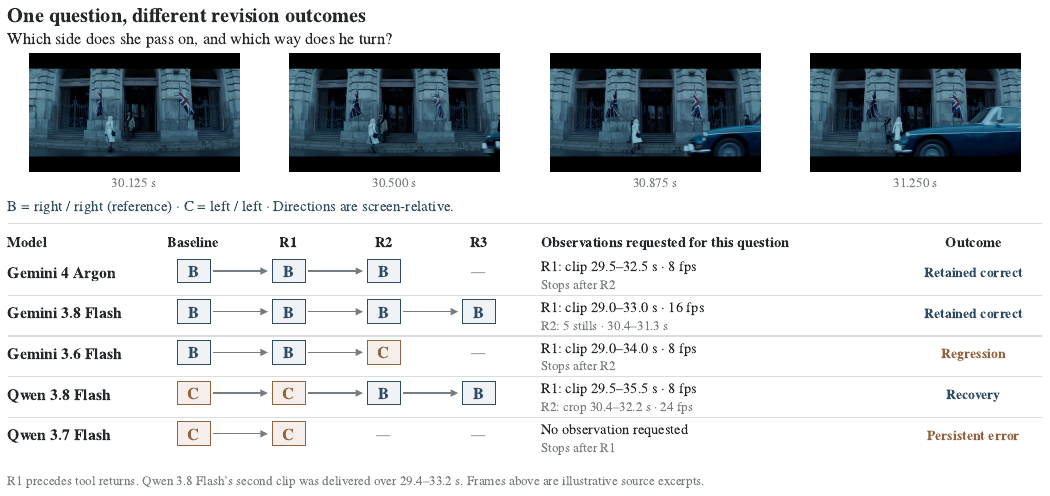}
\caption{\textbf{Opposite revisions on the same question.} Source-frame excerpts and complete answer paths for all five action models. Requests are shown by round; a dash indicates that the session has ended.}
\label{fig:case}
\end{figure}

\paragraph{A shared question, opposite revisions.}
In \cref{fig:case}, Qwen 3.7 Flash stops in round one with an incorrect answer and no observation request. After an 8\,fps clip, Qwen 3.8 Flash changes C to the reference B, retaining it after a 24\,fps crop. Gemini 3.6 Flash instead changes B to C after its 8\,fps clip. Gemini 3.8 Flash and Gemini 4 Argon retain B throughout.

\subsection{Conditions for effective interaction}
\label{sec:exp_variants}
Traces distinguish early stopping from later reversals. Encouraging inspection addresses the former; reliable interpretation is needed for the latter. FastAct makes this distinction concrete: the relevant interval is known, yet models still differ in what they inspect and how they interpret the returned evidence. Task comparisons locate gains; trajectories show how models obtain them, without isolating the causal contributions of observation and reasoning.

\section{Discussion and Conclusion}
\label{sec:conclusion}
Video harnesses make observation part of inference: what a model sees can be shaped by what it has yet to understand. This flexibility can realize capabilities left untapped by fixed observation. Fine-grained actions expose the opportunity: recognizing an event can leave decisive details unresolved. Gains vary across models and tasks: further inspection must address the task's difficulties, and the model must organize observations and use their feedback reliably.

An observation policy should therefore develop with understanding. From a common default, reasoning determines which temporal and spatial details warrant attention. The design challenge is to sustain this reciprocal relation: uncertainty guides evidence acquisition, and the evidence obtained governs whether judgments should change. This coordination must fit the evidence demands of the task and the model's ability to sustain reliable interaction. Useful feedback must therefore inform both the next observation and the decision to retain an existing answer.

\paragraph{Limitations and future directions.}
Our evaluation covers video question answering. The relative gains reported here were measured under a common observation interface and the same three-round limit; alternative tools or stopping rules may change which models benefit most. Controlled evidence interventions could separate visual access from prompting and additional reasoning. Interactive 3D environments, egocentric world modeling, and embodied tasks would test actions that change the environment. Training smaller models on interaction data could directly test evidence seeking, tool use, and answer revision.

\section*{LLM Usage Statement}
LLMs were used in this work only in the following, limited ways. In writing, they served in an assistive capacity: authors drafted the text and used LLMs to edit and refine it. In code and implementation, they likewise assisted with editing and refining author-written code. In data preparation, LLMs were used to generate part of the data (synthetic data, dataset creation, or annotation), and the authors reviewed, edited, and approved all generated output. LLMs were not used for related work and citations, for the design, execution, or analysis of experiments, for figures and visual material, or for research ideas, hypotheses, claims, and conclusions. Neither the full manuscript nor major portions of it were shared with any LLM provider to obtain reviewer-style feedback.

\bibliographystyle{plainnat}
\bibliography{main}
\appendix
\section*{Supplementary Material}

\section{Implementation details}
\label{sec:supp_tools}
\paragraph{Conversation and observation budget.}
The session begins with a global video part, a structured answering/planning prompt, and the questions of the group. Later turns append newly executed observations and a follow-up prompt, retaining conversation history. At most three model rounds and six tool requests per request-producing round are permitted. Requests are ordered by question before execution. Redundant requests and observations exceeding provider limits may be skipped; execution logs distinguish requested from delivered observations. The Gemini path additionally limits the number of video observations in the conversation. These operational restrictions are part of the evaluated harness.

\paragraph{Tools.}
The clip executor caps the requested local frame rate at 24\,fps and can reduce it to satisfy its frame budget; short windows receive boundary padding. The frame tool returns selected stills with a labelled grid, optional crops, and counting markers. Montages order candidate occurrences and add boundary context. Scan requests cover a longer interval at coarse temporal resolution. The overview tool returns a focused textual re-reading of the global video. Motion-energy summaries are automatic attachments for eligible motion questions, not model-requested actions. These tools expose evidence without supplying reference labels.

\paragraph{FastAct source and annotations.}
FastAct is built from high-quality internal video data with human annotation and review. Its 100 source clips contain 1,000 four-option questions, evaluated in 300 question groups. Questions specify the relevant time range and distinguish local action details; the annotated evidence intervals have a median duration of 0.90 seconds. The temporal-demand annotations comprise 693 questions at 8\,fps and 307 at 24\,fps. These strata describe the questions rather than a controlled intervention on the model's input.

\section{Prompt templates}
\label{sec:supp_prompts}
The templates below are extracted from the evaluated prompt builders: \nolinkurl{build_public_test_prompt}, \nolinkurl{build_prompt}, and \nolinkurl{followup_text}. Angle-bracket placeholders denote variable fields or repeated blocks. The question block repeats for each of the $N$ questions; the type and checklist are inferred from the question text. The baseline uses the benchmark's available option letters. No reference answer is included.

\begin{minipage}{\linewidth}
\paragraph{Single-pass baseline.}
The model answers the question group from the default overview.
\begin{lstlisting}[style=prompt]
Target video file: <VIDEO_PATH>

Answer the following <N> question(s) about this video:

Question 1:
<QUESTION_TEXT>
Options:
<OPTIONS>

For each question, set `"answer"` to the single option letter ("A" through "<LAST_OPTION>").
Output your final response as a JSON block with the following exact structure:
```json
{
  "questions": [
    {
      "question": 1,
      "answer": "A",
      "evidence": "brief timestamped justification"
    }
  ]
}
```
\end{lstlisting}
\end{minipage}

\begin{minipage}{\linewidth}
\paragraph{Initial harness turn.}
The first turn adds evidence assessment and observation planning.
\begin{lstlisting}[style=prompt]
You are answering <N> multiple-choice questions about one video. The first video part is the whole video at 1 frame per second, with its audio. This is round 1 of 3: solve what you can from the current observations, and request closer looks (in MM:SS format) for any question that still needs higher frame rate or spatial zoom.

## Question 1 (type: <QUESTION_TYPE>)
<QUESTION_TEXT>
Options:
<OPTIONS>

How to judge question 1:
<QUESTION_CHECKLIST>

<TOOL_DEFINITIONS>

Instructions:
- [ ] For each question, write `evidence` first with exact timestamps in MM:SS.s format, then `status` ("confirmed" or "open"), and `answer`.
- [ ] Mark `status` as "confirmed" when the visual/audio evidence you have seen decides it; otherwise mark it "open", explain what needs checking in `unclear`, and request the right closer look.
- [ ] When new close-look observations are attached for a question, check each option's concrete details against the close looks and update `answer` if a close look supports a different option.
- [ ] Reply with JSON only:
{"questions": [{"question": 1, "evidence": "...", "unclear": "...", "status": "confirmed or open", "answer": "A"}], "requests": [ ... ]}
\end{lstlisting}
\end{minipage}

\paragraph{Available observation requests.}
The following block replaces \texttt{<TOOL\_DEFINITIONS>} in the initial and second-round prompts.
\begin{lstlisting}[style=prompt]
Closer looks you can request in "requests" (up to 6 per round; write times as "MM:SS" or "MM:SS.s", e.g. "01:54" or "04:43.5"):
- {"tool": "clip", "question": k, "start": "MM:SS", "end": "MM:SS", "fps": 4|8|24, "box": [x0, y0, x1, y1] or null, "purpose": "..."}: high-resolution video clip (up to 24 fps). Choose `fps` and window length flexibly (keep each clip within ~60-120 frames: ~4 fps for 15 s, ~8 fps for 10 s, or up to 24 fps for 5 s fast/sub-second actions; "detail" and "motion" are also accepted).
- {"tool": "frames", "question": k, "times": ["MM:SS.s", ...], "cells": ["B2", "B3"] or null, "crop": [x0, y0, x1, y1] or null, "points": [[x, y], ...] or null, "purpose": "..."}: up to 8 full-resolution still frames. Full frames include a 4x4 grid (A1..D4) and 0.1-0.9 coordinate ruler; pass `cells` (e.g. ["B2", "C2"]) or `crop` ([x0, y0, x1, y1]) to zoom into that region at 1536 px; pass `points` ([[x, y], ...]) to draw numbered markers #1..#N for verifying dense single-frame counts.
- {"tool": "montage", "question": k, "times": ["MM:SS.s", ...], "cells": ["B2"] or null, "purpose": "..."}: builds a numbered occurrence ledger montage from candidate event timestamps [t1, t2, ...] plus boundary probe frames [PRE] and [POST] to verify temporal counts without double-counting or missing boundary reps.
- {"tool": "scan", "question": k, "start": "MM:SS", "end": "MM:SS", "purpose": "..."}: 2 fps default resolution for wide ranges (30-120 s).
- {"tool": "overview", "question": k, "purpose": "what to search for across the whole video"}: re-scans the whole video at 1 fps (with audio) if a target event was not found in the local clips, and appends the findings to the overview file.
\end{lstlisting}

\begin{minipage}{\linewidth}
\paragraph{Subsequent turns.}
New observations are appended to the retained conversation. The answer instructions and JSON format from the initial turn are repeated. If a request cannot be executed, the follow-up also reports the request and the reason it was skipped. The final turn omits tool definitions.
\begin{lstlisting}[style=prompt]
The closer looks you requested in your previous reply are attached above, in the order you requested them. This is round 2 of 3. Re-examine every question with everything you have seen in this conversation so far; request further closer looks for any question that still needs them.

<TOOL_DEFINITIONS>
<ANSWER_INSTRUCTIONS_AND_JSON_FORMAT>
\end{lstlisting}
\begin{minipage}{\linewidth}
\begin{lstlisting}[style=prompt]
The closer looks you requested in your previous reply are attached above, in the order you requested them. This is round 3 of 3. This is the final round: give your final answers from everything you have seen in this conversation.

<ANSWER_INSTRUCTIONS_AND_JSON_FORMAT>
\end{lstlisting}
\end{minipage}
For logical question groups in Video-MME-v2, the answering instructions additionally require solving the questions in sequence because group scoring stops at the first incorrect answer.
\end{minipage}

\section{Observation interface in images}
\begin{figure}[H]\centering
\includegraphics[width=\textwidth]{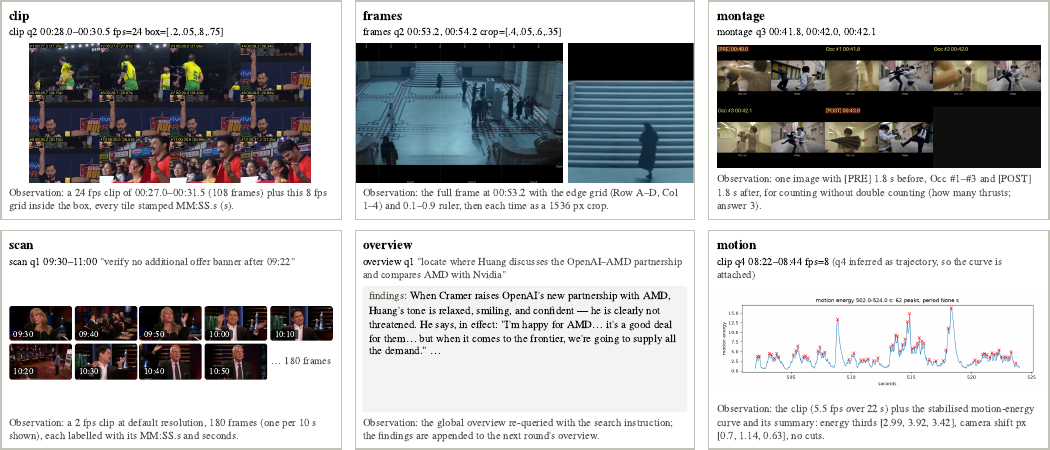}
\caption{\textbf{Actual tool outputs.} Each panel pairs a request with the returned observation; motion summaries attach automatically.}
\label{fig:observations}\end{figure}

\end{document}